\documentclass{article} % For LaTeX2e
\usepackage{iclr2019_conference,times}

\usepackage{amsmath,amsfonts,bm}

\def\eqref#1{equation~\ref{#1}}
\def\1{\bm{1}}

\DeclareMathAlphabet{\mathsfit}{\encodingdefault}{\sfdefault}{m}{sl}
\SetMathAlphabet{\mathsfit}{bold}{\encodingdefault}{\sfdefault}{bx}{n}

\DeclareMathOperator*{\argmin}{arg\,min}

\usepackage{hyperref}
\usepackage{url}
\usepackage{graphicx}
\usepackage{subfig}

\title{One Epoch Is All You Need}

\author{Aran Komatsuzaki \\
School of Mathematics\\
Georgia Institute of Technology\\
Atlanta, GA 30332, USA \\
\texttt{akomatsuzaki3@gatech.edu} \\
}

\iclrfinalcopy

\begin{document}

\maketitle

\begin{abstract}
In unsupervised learning, collecting more data is not always a costly process unlike the training. For example, it is not hard to enlarge the 40GB WebText used for training GPT-2 by modifying its sampling methodology considering how many webpages there are in the Internet. On the other hand, given that training on this dataset already costs tens of thousands of dollars, training on a larger dataset naively is not cost-wise feasible. In this paper, we suggest to train on a larger dataset for only one epoch unlike the current practice, in which the unsupervised models are trained for from tens to hundreds of epochs. Furthermore, we suggest to adjust the model size and the number of iterations to be performed appropriately. We show that the performance of Transformer language model becomes dramatically improved in this way, especially if the original number of epochs is greater. For example, by replacing the training for 10 epochs with the one epoch training, this translates to 1.9-3.3x speedup in wall-clock time in our settings and more if the original number of epochs is greater. Under one epoch training, no overfitting occurs, and regularization method does nothing but slows down the training. Also, the curve of test loss over iterations follows power-law extensively. We compare the wall-clock time of the training of models with different parameter budget under one epoch training, and we show that size/iteration adjustment based on our proposed heuristics leads to 1-2.7x speedup in our cases. With the two methods combined, we achieve 3.3-5.1x speedup. Finally, we speculate various implications of one epoch training and size/iteration adjustment. In particular, based on our analysis we believe that we can reduce the cost to train the state-of-the-art models as BERT and GPT-2 dramatically, maybe even by the factor of 10. 
\end{abstract}

\section{Introduction}
Recently, unsupervised models have been growing rapidly and becoming more promising than previously expected. 
Notable examples include GPT-2, BERT and Sparse Transformer. The performance of these models can be scaled up stably if they are given more parameters and more training data. However, there is still a visible gap between their performance, especially that of generative models, and that of human. From the current trend, it seems that the gap can be filled up if we can keep scaling up and developing architectures for better long-range coherence such as Transformer-XL \citep{xl} and Sparse Transformer \citep{sparse}. Since the training of the best generative models already costs tens of thousands of dollars, naively scaling up is not a practical option. 

In fact, there is one obvious feature of the currently dominant practice in machine learning that can be modified for significant performance gain. This is to train for multiple epochs. While multi-epoch training is reasonable in data-scarce settings such as supervised classification, this turns out to be inefficient for data-abundant unsupervised learning according to our results. This may be a trivial observation, but many recent papers have used multi-epoch training as shown in Table \ref{epochs}. While many papers do not report the number of epochs, it is reasonable to assume that the number is between 10 and 200. Note that GPT, BERT and GPT-2 were trained on an original dataset created by their authors. Hence, they could have increased the dataset size and reduced the number of epochs for better performance. On the other hand, many other papers have trained their model on a standard dataset for many epochs to make the comparison with the previous state-of-the-art fair. Also, there are many papers that do not report the number of epochs used for the training. Since the total computational resources spent on the training is proportional to not only the number of parameters but also the number of epochs, the current practice should be reconsidered. For example, we need to create larger standard datasets, and the models have to be trained for only one epoch for a fair comparison.  

\begin{table}[t]
\caption{The number of epochs used for the training.}
\label{epochs}
\begin{center}
\begin{tabular}{ll}
\multicolumn{1}{c}{\bf Model}  &\multicolumn{1}{c}{\bf Epochs}
\\ \hline \\
GPT \citep{gpt}   & 100 \\
SPN \citep{spn} & Not reported \\
BERT \citep{bert} & 40\\
Mesh Transformer \citep{mesh}         & 10 \\
Transformer-XL \citep{xl}             & Not reported\\
GPT-2 \citep{gpt2} & Not reported (20 or 100)\\
Sparse Transformer \citep{sparse} & 70 - 120\\
\end{tabular}
\end{center}
\end{table}

\section{Related Works}
While we are not aware of any work that investigates training for one epoch with enlarged dataset, there are many works that suggest the training on larger dataset (with a large number of epochs) to improve the performance. Notably, the result of GPT \citep{gpt}, BERT \citep{bert} and GPT-2 \citep{gpt2} implies that the training on a large dataset leads to a significant improvement in performance. 

Some works have analyzed the relationship between the performance and the dataset size \citep{unreasonable, scaling}. Our work is most closely related to \citet{scaling}, which showed that by training a language model on the subsets of 1 Billion Word Language Model Benchmark (LM1B) \citep{1b} of the varying size the performance and the dataset size follow a robust power law given a sufficient parameter budget. This comparison is more precise than the aforementioned work. The difference from our work is that as the dataset size increases, they also let the model to consume more computational resources by increasing the parameter budget and fixing the number of epochs. Also, they do not investigate the trade-off between the parameter budget and the number of iterations unlike our work. This means that the performance improvement promised by their result can be achieved only if computational resources are increased. In our work, we achieve to improve the performance without increasing the computational resources.

One of the experiments of \citet{tips} shows that larger neural machine translation (NMT) dataset results in higher BLEU with the same parameter budget and the same number of iterations, i.e., with smaller number of epochs. Since it is generally not easy to enlarge a NMT dataset, the training with a small number of epochs was not investigated in the paper.  
\section{Methods}
\subsection{One Epoch Training and Size/Iteration Adjustment}
\subsubsection{To See Performance Improvement over Conventional Setting}
First, we describe one epoch training conversion procedure, which converts a conventional multi-epoch training into a more efficient one epoch training. 
\begin{enumerate}
    \item The dataset size is increased (e.g. by sampling from Internet a la WebText), so that, while training for the same number of iterations as before, the same sample is never reused.
    \item Any regularization method is eliminated. 
\end{enumerate}
This process substantially improves the performance with the same computation cost unless the model size is much smaller compared with the dataset size, in which case the improvement is less. 

Then, in order to further improve the performance without increasing the computation cost, we adjust the model size and the number of iterations to be performed while keeping their product constant. We find that, if the ratio of the number of tokens in the dataset over the number of parameters of the model is closer to 5, this is likely to give the optimal performance under one epoch training given the cost constraint. We denote the (initial) number of parameters by $P$ (or $P_0$) and the (initial) number of tokens to be processed by $T$ (or $T_0$). Note that $T=cI$, where $c$ is the number of tokens per minibatch and $I$ is the number of iterations.
\begin{enumerate}
  \setcounter{enumi}{2}
  \item We set $P$ and $T$ according to some heuristics. For example, we can perform this by setting the ratio $T/P$ as close to 5 as possible while keeping their product constant, or equivalently by solving the following: \[\argmin_{P,T}|\log(5)-\log(T/P)|\quad\quad\text{subject to}\quad\quad PT=P_0T_0.\] 
\end{enumerate} In practice, the range of $P$ has only a small number of elements, since more choices do not usually result in a significant speedup. Therefore, finding $P$ and $T$ is quite simple. 
\subsubsection{How to Use Them in Practice}
The above operations are useful to see performance improvement with one epoch training and model size adjustment over the original multi-epoch setting. In practice, one is more interested in how to use these techniques in practice, rather than starting from a given multi-epoch training prototype. From our argument so far, this is quite simple to do. 
\begin{enumerate}
  \item Choose the number of iterations $I$.
\end{enumerate}
Note that the total computation cost scales up quadratically with respect to $I$, since the optimal model size scales up linearly with respect to $I$. From this and the available computation budget, one can choose the value of $I$. The optimal model size is then chosen as in the above method. Let us recall that we have $T=cI$. We use the same notations as before. 
\begin{enumerate}
  \setcounter{enumi}{1}
  \item We set $P$ according to some heuristics. For example, we can perform this by setting the ratio $T/P$ as close to 5 as possible while keeping their product constant, or equivalently by solving the following: \[\argmin_{P}|\log(5)-\log(T/P)|.\] 
  \item The model with size $P$ is trained for $I$ iterations for one epoch without any regularization method. 
\end{enumerate}

\subsection{Justifications for One Epoch Training}
The justifications are quite intuitive. First, note that one epoch training substantially improves the diversity of the samples processed by the model over the course of training. Training for $E$ epochs is roughly equivalent to training on a shuffled dataset consisting of $E$ copies of the original dataset for one epoch. This means that the diversity of the original dataset is $E$ times less than that of the  one epoch training. Greater dataset size also implies greater diversity, and both of these are known to improve the performance \citep{scaling, gpt2}. For example, WebText led to a better generation quality than LM1B not only due to the greater dataset size and larger context size but also due to the diversity of the dataset. 

Also, under one epoch training, sampling from the training data distribution is practically identical to sampling from the underlying data manifold (and therefore from that of the test dataset). This is unlike the multi-epoch training, since sampling each sample again cannot occur when one samples from the data manifold, whose cardinality is practically infinite. This sampling discrepancy is a primary cause of overfitting, which is a well-known fact, and the lack of overfitting observed in one epoch training supports this. Overfitting is usually exacerbated as the number of iterations (and hence the number of epochs) increases. This leads to better speedup with one epoch training when the number of epochs of the original multi-epoch training is larger. Finally, note that averaging the train loss per minibatch measured on the past $n$ minibatches is approximately equal to the test loss if $n$ is small enough. Hence, validation with test/validation dataset is not crucial.

\section{Experiments}
Unless specified otherwise, the hyperparameters are as described here. We train base Transformer decoder \citep{transformer} with some modifications (as described below) for language model. The dataset used is 1 Billion Word Language Model Benchmark (LM1B). We do not use any regularization method unless specified otherwise. Whenever the number of parameters is given below, it contains the number of parameters of softmax and the embedding. As in \citet{bert, gpt2}, we have $d_{ff}=4d_{model}$, $d_q=d_k=d_v=d_{model}$ and $h=\frac{d_{model}}{64}$, where $h$ denotes the number of attention heads. Whenever we use the symbol $d$ below, it stands for $d_{model}$. More details on the dataset and hyperparameters are provided in Appendix. 

We are going to use the subsets of LM1B with varying size as in \citet{scaling}. Note that the test loss achieved in our experiments are inferior to the models of \citet{scaling} and the state-of-the-art models for at least one of the following reasons: (1) the number of trained iterations and/or the model size is significantly smaller in our case, and (2) the vocabulary size was set 10,000 in \citet{scaling} to save the computation (unlike about 800,000 in the most papers, including ours), which means their loss values are significantly smaller than what they would be if the vocabulary size was the same as ours. 

This experiment verifies that the performance of a language model is improved by the one epoch training, and regularization harms the training under the one epoch training, i.e., if no sample is reused in the training. We denote "training with one epoch training" by $S$, "training for multiple epochs" by $M$ and "using dropout" by $D$. For example, if a model is trained with single epoch training and $p=0.1$, we denote it by $SD$. We train the Transformer for the four cases: $S$, $M$, $SD$ and $MD$. 

We set $d_{model}=512$ and train for 65,000 iterations. The number of parameters of the model is 45M. We also set $p=0.1$ for the cases in which dropout is used, unless specified otherwise. The dataset size is 45M tokens (processed in 6,500 iterations) for the multi-epoch case and 450M tokens for the single epoch case. Hence, 10 epochs are performed in total for the former case. This choice of dataset size is due to the popular custom of training a language model whose size is close to the number of tokens of the dataset. The result is shown in the top left figure of Fig. \ref{epoch}. 

The other cases in the figure are provided to demonstrate how the magnitude of speedup differs depending on whether the model is more overparametrized or underparametrized. For example, the Right case has a smaller model size (i.e. more underparametrized). Likewise, the Bottom case is more overparametrized. The speedup (1.9x) of Right is smaller than the speedup (3.3x) of Left and Bottom. Thus, the speedup is smaller if the model is more underparametrized, which is expected. For the Right case and the Bottom case, only the best-performing dropout probability is shown (e.g. $p=0$ for the Right case and $p=0.1$ for the Bottom case). In any case, the model performs worse if $p>0.1$.  

\subsection{one epoch training}
\begin{figure}
    \centering
    \subfloat{{\includegraphics[width=0.5\linewidth]{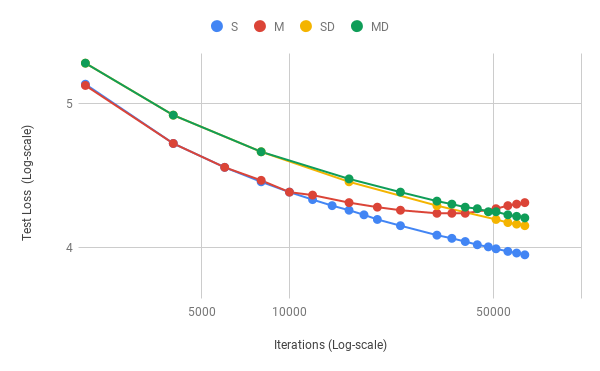} }}%
    \subfloat{{\includegraphics[width=0.5\linewidth]{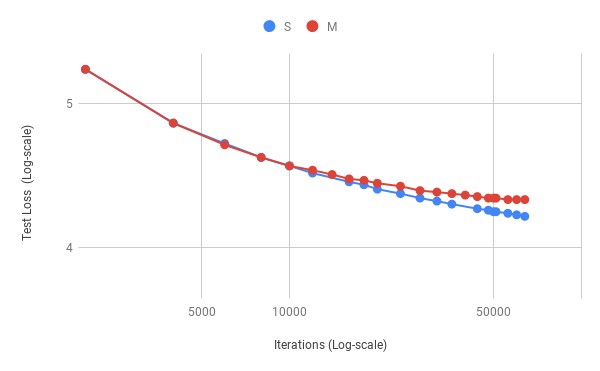} }}\\
    \subfloat{{\includegraphics[width=0.5\linewidth]{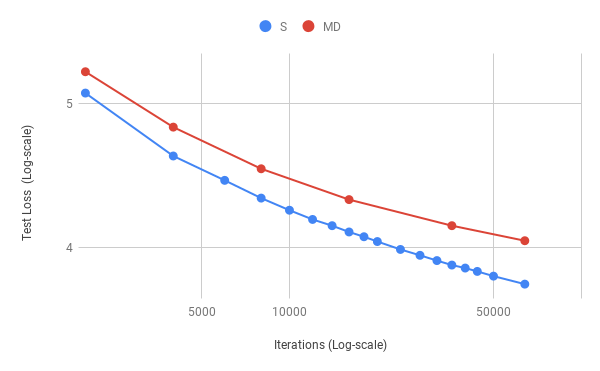} }}%
    \caption{Learning curve of LM for 65,000 iterations on subsets of LM1B with different configurations. }
    \label{epoch}
\end{figure}

Here, the speedup is calculated as follows. First, we compute the number of iterations for $M$ or $MD$ to achieve the best loss. Also, we compute the number of iterations for $S$ to achieve the best loss achieved by $M$ and $MD$. The speedup is defined to be the ratio of the former quantity over the latter. For the case of the left figure of Fig. \ref{epoch}, the former is 65,000, whereas the latter is 20,000. Thus, the speedup is $\frac{65000}{20000}\approx 3.3$ times. In the table, "$E=10$" refers to the speedup for the first 10 epochs are trained, whereas "$E=5$" refers to the speedup for the first 5 epochs are trained. The speedup for the former case is just as explained above, and the speedup for the latter case can be calculated likewise by ignoring the data on the plot of Fig. \ref{epoch} after the first 5 epochs. The result of Table \ref{config} suggests that the speedup ($E=10$) is greater than the speedup ($E=5$). This implies that the speedup is greater if the number of epochs of the original multi-epoch training is greater. 

When dropout is used, the curve is shifted upward. The magnitude of shift increases as $p$ increases or as the regularization becomes stronger, which slows down the training. Furthermore, the left figure of Fig. \ref{epoch} suggests that, if dropout is used, the speed of the training does not change much whether one epoch training or multi-epoch training is used. This suggests that each sample cannot be memorized well under dropout, which is well-known. If the availability of data is limited (e.g. in supervised learning) and if unsupervised pretraining does not help, one can attempt to mitigate the gap between $S$ and $MD$ with the adaptive dropout as described in Appendix. Note that this adaptive dropout is much more inefficient than one epoch training if the data is plentiful, which is our assumption. 
\begin{table}
  \centering
\begin{tabular}{ |c|c|c|c|c|c|c| } 
\hline
& $d$ & Parameters & Epochs & Iters./Epoch & Speedup ($E=10$) & Speedup ($E=5$)\\
\hline
Left & 512 & 45M & 10 & 6500 & 3.3 & 1.8\\
Right & 256 & 18M & 10 & 6500 & 1.9 & 1.5\\
Bottom & 1024 & 128M & 10 & 6500 & 3.3 & 2.6\\
\hline
\end{tabular}
\caption{Configuration of each figure of Fig. \ref{epoch}.}
\label{config}
\end{table} 
\subsection{Power law}
Under the one epoch training, analyzing the training becomes simpler, since regularization does not need to be taken into consideration. 

\begin{figure}
    \centering
    \subfloat{{\includegraphics[width=0.5\linewidth]{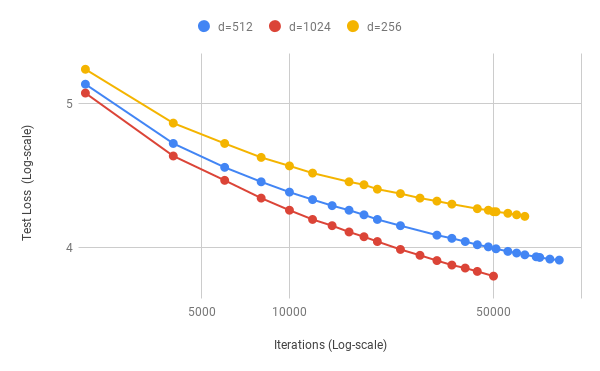} }}%
    \subfloat{{\includegraphics[width=0.5\linewidth]{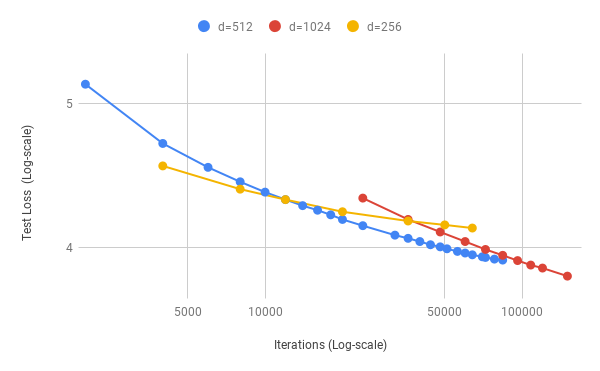} }}%
    \caption{(Left) Log-log plot of learning curve over iterations. (Right) Log-log plot of the learning curve scaled according to the per-iteration FLOPS with respect to the $d=512$ curve, which is fixed at its original position as with the scaling of the x-axis.}
    \label{speed}
\end{figure}

\begin{figure}
    \centering
    \subfloat{{\includegraphics[width=0.5\linewidth]{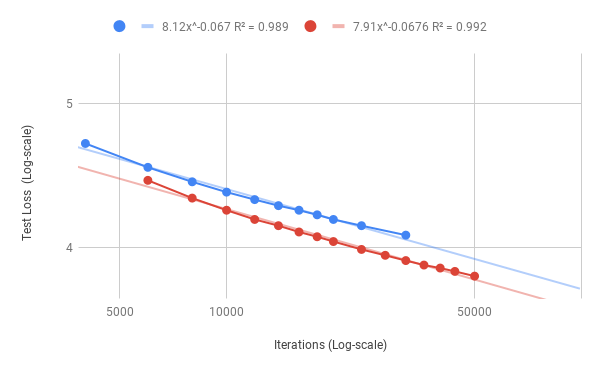} }}%
    \subfloat{{\includegraphics[width=0.5\linewidth]{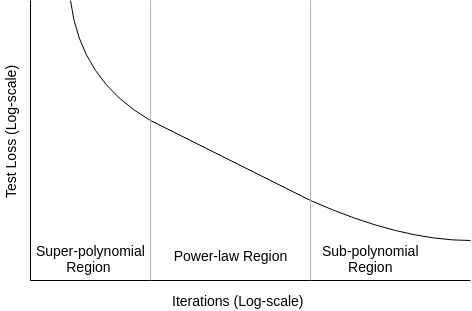} }}%
    \caption{(Left) Log-log plot of partial learning curve of LM over iterations with a line fit. (Right) Sketch of learning curve over iterations.}
    \label{scaling2}
\end{figure}

The left figure of Fig. \ref{speed} shows the log-log plot of the curve of test loss over the iterations for different widths. Each curve has a structure as depicted in the right figure of Fig. \ref{scaling2}. Observe that the curve first enters a super-polynomial region, where the loss decreases faster than any polynomial, since the parameters are not yet saturated with many training samples. Then, the curve enters a linear region, which is, in fact, a power-law region, since the plot is log-log. This means on this region the test loss follows $\ell = ax^{-k}$, where $x$ is the number of iterations and $a$ and $k$ are some constant. As the parameters are oversaturated with the training samples, the loss stagnates, and the curve enters sub-polynomial region convergent to a constant. As the parameter budget increases, the super-polynomial region and the power-law region expands, which contributes to the superior performance of larger models. Unlike the multi-epoch setting, the loss decreases more steeply, monotonically and for longer iterations. While the gap of loss among each model is small at the beginning, it increases as the iteration increases.  

\begin{figure}
    \centering
    \subfloat{{\includegraphics[width=0.5\linewidth]{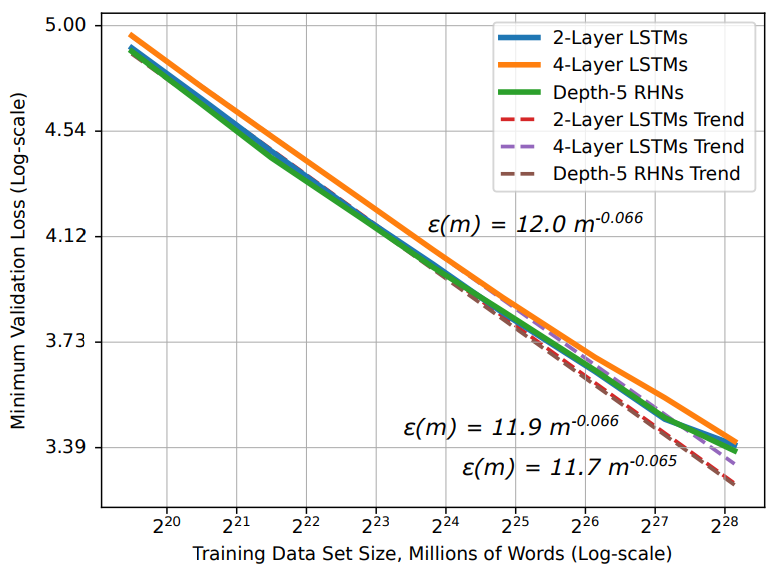} }}%
    \subfloat{{\includegraphics[width=0.5\linewidth]{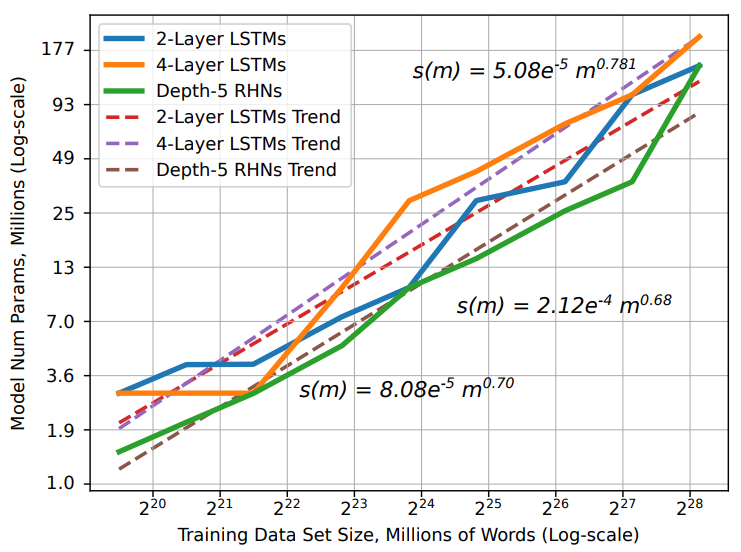} }}%
    \caption{Learning curve of LM on subsets of LM1B with varying size (cited from \citet{scaling}).}
    \label{scaling}
\end{figure}

The left figure of Fig. \ref{scaling2} shows the line fit for the power law region of each model. Notably, the power law exponent (about $-0.067$) is approximately equal to the power law exponent of found in \citet{scaling}, which is shown in the left figure of Fig. \ref{scaling} for convenience. The major difference between their experiment and ours is that they trained a model with different parameter budget sufficiently large for each dataset size for many epochs until the best loss is achieved. The right figure of Fig. \ref{scaling} shows the parameter budget for each dataset size.     

\subsection{Size/Iteration Adjustment}
As far as large-scale Transformer language model is concerned, modifying the architecture with a method such as neural architecture search leads to a smaller gain compared with scaling up the model with more data or parameter. Hence, it suffices to consider changing depth and width only. For simplicity, we consider changing only width. Let us say we train a model with the number of parameters $P$ for $I$ iterations. Then, the total FLOPS of the training is proportional to $PI$, assuming that the GPU is used efficiently, which is easy at large-scale. We are interested in finding the range of optimal $I$ for a given number of parameters (or more conveniently, width). For this, we remap the curves in the left figure of Fig. \ref{speed} to adjust for the difference in per-iteration FLOPS and derive the range of optimal number of iterations for a given model size as described in Appendix, which is shown in the right figure of Fig \ref{speed} and Table \ref{optimal}, respectively. The range of optimal number of iterations is then converted to the number of tokens processed and divided by the number of parameters of the model, which is shown in the rightmost column of Table \ref{optimal}. By taking the intersection of the ranges, we obtain $[1.8,11.5]$. Since the geometric mean of the boundary values is $\sqrt{1.8\times11.5}\approx 5$, in our heuristic for adjustment we try to make the ratio of the number of processed tokens over the number of parameters, or $T/P$, as close to 5 as possible. This result suggests that five words per parameter can be the most efficiently compressed, at least in our setting. 

\begin{table}
  \centering
\begin{tabular}{ |c|c|c|c|c|c|c| } 
\hline
$d$ & Parameters & Optimal Iters. & (Optimal Tokens)/Params.\\
\hline
256 & 18M & $[0, 30000]$ & $[0,11.5]$\\
512 & 45M & $[12000, 84000]$ & $[1.8,12.9]$\\
1024 & 128M & $[28000,\infty)$ & $[1.5,\infty)$\\
\hline
\end{tabular}
\caption{Optimal number of iterations and ratio.}
\label{optimal}
\end{table} 

\begin{table}
  \centering
\begin{tabular}{ |c|c|c|c|c|c|c| } 
\hline
& old $d$ & new $d$ & Speedup & Combined Speedup\\
\hline
Left & 512 & 512 & 1 & $3.3\times 1=3.3$\\
Right & 256 & 512 & 2.7 & $1.9\times 2.7\approx5.1$\\
Bottom & 1024 & 512 & 1.3 & $3.3\times 1.3\approx4.3$\\
\hline
\end{tabular}
\caption{Speedup with size/iteration adjustment and total speedup.}
\label{total}
\end{table} 

Since we have $I=65000$, the optimal width among $\{256, 512, 1024\}$ is $d=512$, which follows from either our heuristics or the right figure of Fig. \ref{speed}. Each configuration undergoes a speedup due to changing the width to $d=512$. Table \ref{total} summarizes the speedup achieved with size/iteration adjustment and the total speedup combined with that of one epoch training. The result agrees with our intuition that size/iteration adjustment would lead to a better speedup if the original size/iteration proportion is more skewed. In terms of the combined speedup, this also agrees with this same intuition.
\section{Implications and Remarks}
\subsection{state-of-the-art models are likely to undergo better speedup}
Table \ref{sota} shows the number of epochs and the ratio of the tokens processed over the number of parameters of the state-of-the-art models. The original multi-epoch training is first converted to the one epoch counterpart, which results in an increase of the dataset size by the factor of the number of epochs. This contributes to the ratio being substantially larger than 5. Combined with the number of epochs being far greater than 10, it is reasonable to expect that we can accelerate the training of the state-of-the-art models with the factor substantially greater than our 3.3-5.1x speedup, maybe even by the factor of 10.  

\begin{table}[t]
\caption{The number of epochs used for the training.}
\label{sota}
\begin{center}
\begin{tabular}{lll}
\multicolumn{1}{c}{\bf Model}  &\multicolumn{1}{c}{\bf Epochs} &\multicolumn{1}{c}{\bf Tokens/Params.}
\\ \hline \\
BERT  & 40 & 340\\
Mesh Transformer  & 10 & 1.6-57\\
GPT-2  & 20 (or 100) & 106 (or 530)\\
\end{tabular}
\end{center}
\end{table}
\subsection{Range of applicability}
One epoch training and size/iteration adjustment are also applicable to many unsupervised learning algorithms on data of any modality as well as semi-supervised learning as \citep{semi}. Regularization methods are crucial for and more dominant in many computer vision tasks. Hence, they may benefit from one epoch training even better. It is very likely that our results hold for larger-scale models due to the nature of our methods. We should perform more comprehensive studies to measure how much speedup is achieved in each setting and refine our heuristics for size/iteration adjustment.  
\subsection{EfficientNet Scaling with the number of iterations}
\citet{efficient} showed that, for image classification, scaling up a model by jointly searching for the scaling factor of each scaling component (e.g. depth) with grid search leads to a dramatic improvement over scaling up one or two components only without extensive search. In our case, there are three scaling factors: depth, width, the number of iterations. It is reasonable to expect that this will give more favorable scaling than heuristics. 
\subsection{Caveats on fine-tuning} 
One notable usage of unsupervised models such as GPT-2 and BERT is fine-tuning the pre-trained model to a small specialized dataset. Since we suggest that no regularization method is used for the training of the model, one would suspect that the lack of regularization would cause overfitting during the fine-tuning process. We argue this may not be necessarily the case. Note that GPT-2 does not use any regularization method, and \citet{bert} suggests to fine-tune BERT for only a few epochs. It is also important to note that one-epoch training improves the performance of the pre-trained model and therefore requires a smaller number of iterations on the fine-tuning dataset to reach to the same performance. 
\subsection{Sample efficiency of left-to-right language model and BERT}
BERT is known to be more sample efficient than other language models such as the left-to-right language model as shown in papers as \citet{bert}. We believe that one of the reasons is that, for BERT, the mask (hence the input and the target) of a sample is different for each epoch. Under the one epoch training, this advantage vanishes. Hence, we believe it diminishes the efficiency gap between BERT and the left-to-right language model. If the efficiency gap is small, the latter model is far more preferable than BERT. For example, the text generation capability of BERT is poorer than its left-to-right counterpart. Hence, it requires the softmax to be trained from scratch for each task upon the fine-tuning, possibly different softmax for each task. On the other hand, left-to-right language models can, in principle, perform, with not necessarily better performance, any task BERT can perform without fine-tuning, even with zero-shot learning. For example, GPT-2 can perform text classification by not only predicting the label (e.g. "Math") but also by generating a text (e.g. "It's about math."). This adds far more flexibility to GPT-2, and GPT-2 can be trained with various tasks seamlessly. By exploiting this fact, one can combine various notable (training) datasets to the general-purpose dataset as WebText for training. After the training, without fine-tuning the model may perform well on their test datasets. The performance may be improved by instead adding a few copies of the added datasets instead of just one. Also, note that, while fine-tuning performs well in general, it has been scarcely investigated on whether training on narrower distribution suddenly right after the pretraining would lead to something similar to catastrophic forgetting. After the pretraining is finished, it may be more effective to simply mix the samples of the task of the interest to the samples from the pretraining dataset to continue the training without performing the conventional fine-tuning. 

As it was mentioned in \citet{gpt} that fine-tuning of GPT converges within a few epochs, it is reasonable to assume the same for GPT-2. Given the strong influence of the pretraining method on the final performance on the takes Under the one epoch training, GPT-2 may not need fine-tuning process if the task is known during the pre-training process, as described below.

%Rather than using fine-tuning for GPT-2, it may be better to throw all the available training datasets of various down-streaming tasks to WebText, which seems more efficient. The label can be converted to natural language, so a separate softmax for the labels is unnecessary.   ........................
\subsection{Shift of attention from regularization to model capacity}
Deep learning research has almost always been performed under multi-epoch training with regularization method, so the resulting generalization was more important than improving the actual capacity of the model. It has been believed that large-scale training is the only viable way to measure the actual model capacity, as the regularization is not needed in this setting. This has limited the research into improving the actual model capacity, which can be afforded only by affluent research groups, and proliferated the works on improving the regularization methods. However, one epoch training may shift the attention to the improvement of model capacity. The methods prone to overfitting and regarded not viable may be reexamined. Some promising ideas that tend to overfit include mixture-of-experts \citep{moe} and optimization methods exploiting the second order information such as K-FAC \citep{kfac}. 
\subsection{Creation of New Datasets and Comparison of Models} 
As mentioned above, we should create new standard datasets on which a model is trained for one epoch only. We consider the case of language model, but a similar argument holds for other tasks and the data of other modality. A good candidate is a subset of a dataset similar to WebText. Since one can compare model improvement more or less regardless of the model size due to the lack of regularization, subset larger than 400M tokens may not be always necessary. Note that the total computation cost scales quadratically with respect to the dataset size. Since the model has to be trained for one epoch only, substantially larger datasets can be also used for a small amount of computation if necessary. There are two ways to compare models. 

1) The dataset creator first makes a plot similar to the right figure of Fig. \ref{speed} using a subset with, say, 400M tokens and identifies the optimal model sizes on the iterations corresponding to, say, 100M tokens, 200M tokens and 400M tokens. For example, in our experiment the optimal model size on 100M tokens would have the width of 512. Then, the model designer would compare the performance of the models with this optimal model size according to their performance after they are trained on the dataset of the chosen size for one epoch. 

2) This method is more precise than the above method. The model designer makes a plot similar to the right figure of Fig. \ref{speed} using a subset with 400M tokens and his proposed model. The plot is then combined with the plot of the state-of-the-art model by adjusting for the difference in per-iteration FLOPS.  
\subsection{Data Augmention with Internet} 
One can also exploit the Internet by augmenting the dataset for the task of the interest by searching for data relevant to the task and adding it to the dataset. Often, poor performance has been attributed to architecture or optimization. However, the actual cause is often because the dataset does not contain a sufficient amount of information useful for the task of the interest. If the model's poor performance is alleviated by this data augmentation, it is likely due to none other than the insufficient information. This is inevitable, since, for example, if the training dataset consists of randomly sampled webpages, the distribution of data on Internet does not necessarily align with what a person would have processed in his life. This mismatch would result in the sample complexity of the language model being poorer than that of human, which is permissible to a certain extent but can be improved by a mismatch-aware dataset sampling strategy.   
\subsection{On sampling data from Internet} 
WebText consists of the webpages that are sampled by selecting the links posted to Reddit with more than 3 karmas. In order to increases the number of webpages by more than several order of magnitude, the strategy of sampling should be more lenient to the corrupt webpages. \citet{common} pointed out that CommonCrawl contains a large portion of corrupt samples, which makes it unsuitable for the training. The proportion of the corrupt samples in CommonCrawl is substantially higher than 50\%, hence most of the resources are spent on training on the useless corrupt samples. One can study how the performance depends on the proportion of corrupt samples in the dataset. From this, one can estimate the upper bound of the proportion of corrupt samples such that the performance degradation is not significant. For example, if the proportion can be merely 10\%, the waste of resources is rather negligible. 

Having some corrupt samples in the dataset to some extent may be beneficial for the model to distinguish the corrupt webpages from the non-corrupt ones. Within the dataset containing the webpages of both types, the distribution of the corrupt samples must be easily separated from that of the non-corrupt samples from the perspective of the trained model. When the dataset contains some corrupt samples, one can prevent them from generating the output that resembles the corrupt samples as follows. For example, note that it is very unlikely that corrupt text is followed by non-corrupt text in the same webpage, or vice versa. Hence, it is interesting to check whether a model conditioned on a non-corrupt text almost always generates non-corrupt text. If a model happens to start generating a corrupt text either conditionally or unconditionally, this may be detected by the perplexity of the generated output. Corrupt text tends to have the perplexity too high or too low. This fact may be also exploited at the training stage. For example, the samples with unusual perplexity can be discarded from the training dataset.   

The sampling method of WebText can be easily expanded or generalized. In addition to using the karmas of Reddit and its variants on other webpages, we can utilize the meta data of each webpage and statistical analysis of the content. For example, non-corrupt webpages must have non-negligible amount of traffic from human users rather than bots, and they must have a clearly distinct access pattern. Also, the distribution of characters or the most frequent words of the corrupt webpages may be different from that of the non-corrupt webpages (e.g. Zipf's law). 

\section{Conclusion}
The following summarizes our work:
\begin{itemize}
    \item The conventional multi-epoch training can be improved by enlarging the dataset, adjusting the model size and the number of iterations appropriately and training for one epoch only. A heuristics for the adjustment was devised.
    \item Overfitting does not occur in one epoch training, and regularization does nothing but slows down the training.
    \item The loss curve over the iterations for a given model size follows power-law rather extensively.
    \item Based on our analysis, we believe we can reduce the cost to train the state-of-the-art models as BERT and GPT-2 dramatically, maybe even by the factor of 10.
    \item One epoch training and size/iteration adjustment are promising for not only language model but also many other unsupervised or semi-supervised learning tasks on data of any modality. 
    \item We can possibly efficiently scale up a model using an analog to EfficientNet scaling with not only the conventional scaling factors such as depth and width but also the number of iterations as the scaling factors. 
    \item The sample efficiency gap between BERT and left-to-right language model is likely to diminish with one epoch training. GPT-2 may replace BERT due to its flexibility.  
    \item Since the overfitting does not occur with one epoch training, more attention will be paid to improving model capacity, which becomes easier to observe. The methods prone to overfitting and regarded not viable may be reexamined.
    \item We should create new standard datasets to evaluate and compare newly proposed models under one epoch training and size/iteration adjustment. We have provided two possible evaluation methods with such datasets. 
    \item We discuss about data augmentation with Internet and how to efficiently sample data from Internet to expand the training dataset. 
\end{itemize}
Future works will hopefully verify our results and claims on larger-scale and on other kinds of tasks. They can also continue to explore for further implications of one epoch training and model size adjustment, as they have been scarcely investigated.
\subsubsection*{Acknowledgments}
We are grateful to Lukasz Kaiser and Isaac Poulton for his valuable feedback on our work.
\bibliography{iclr2019_conference}
\bibliographystyle{iclr2019_conference}
\newpage
\section{Appendix}
\subsection{Further Details on dataset and hyperparameters}
LM1B consists of 28 million sentences (training dataset) from news articles with about 10 thousand sentences for test dataset \citep{1b}. Each sentence is curtailed to the first 50 words. Each iteration consists of a minibatch of 256 sentences, and the average length of a sentence is 27. We set the number of layers to 6. We use word-level tokens and (not tied) adaptive input and softmax \citep{adap}. The cutoff for adaptive softmax/input is $[4000, 20000, 100000]$ to save the memory budget and computation cost at the cost of slightly degraded performance. We do not use checkpoint averaging or label smoothing. We use PyTorch with a single V100 GPU and mixed precision training. Unlike the Transformer of \citet{transformer}, LayerNorm is placed before the self-attention module and ReLU in our case as in \citet{sparse}.
\subsection{Adaptive dropout}
If the availability of data is limited (e.g. in supervised learning) and if unsupervised pretraining does not help, one can attempt to mitigate this problem with the adaptive dropout as described below. The dropout probability is a monotonically increasing function of the number of epochs trained. In particular, it is set zero for the first epoch. It is likely that the gap between one epoch training and multi-epoch training (see the gap in the left figure of Fig. \ref{epoch}) with this dropout is smaller. However, based on the trend, the gap is still likely to increase as the number of epochs increases. Thus, this method is much more inefficient than one epoch training if the data is plentiful. Since the assumed setting is very rare, we do not investigate this direction further. 
\subsection{Further Details on Fig. \ref{speed}}
In this section, we discuss how to convert the left figure of Fig. \ref{speed} to the right figure and how to obtain the range of the optimal number of iterations for each model size. The curve for $d=512$ is fixed at the same position, but other curves are moved to take into account the difference in per-iteration FLOPS. The per-iteration FLOPS of $d=1024$ and $d=256$ is 3 times larger and 2.5 times smaller than that of $d=512$, respectively. Hence, the curve of $d=1024$ is moved left by the factor of 3. A similar operation is performed on the curve of $d=256$. The $d=256$ curve and the $d=512$ curve intersects at 12,000 iterations, and the $d=512$ curve and the $d=1024$ curve intersects at 84,000 iterations. This means that if the model with $d=512$ is used, the number of iterations should be greater than 12,000 and less than 84,000 to minimize the computation cost. On the other hand, if the model with $d=256$ is used, the number of iterations should be less than $12000\times 2.5=30000$. Likewise, if the model with $d=1024$ is used, the number of iterations should be greater than $\frac{84000}{3}=28000$ and less than a certain upper bound.  
\end{document}